\relax
\documentclass[11pt]{article}
\usepackage{fullpage}
\usepackage{times}  
\usepackage{helvet}  
\usepackage{courier}  
\usepackage{url}  
\usepackage{graphicx}  

\usepackage[utf8]{inputenc} 
\usepackage[T1]{fontenc}    
\usepackage{url}            
\usepackage{booktabs}       
\usepackage{amsfonts}       
\usepackage{nicefrac}       
\usepackage{microtype}      
\usepackage{color}

\usepackage{graphicx} 
\usepackage{subfigure}

\usepackage{algorithm}
\usepackage{algorithmic}
\usepackage{amssymb}
\usepackage{amsthm}
\usepackage{amsmath,epsfig}
\usepackage{epstopdf}

\usepackage{enumitem}
\newif\ifsubmit

\submitfalse

\ifsubmit
\newcommand{\bo}[1]{}
\else
\newcommand{\bo}[1]{\textcolor{blue}{Bo: #1}}
\fi

\begin{document}

\title{DGCNN: Disordered Graph Convolutional Neural Network Based on the Gaussian Mixture Model}
\author{
Bo Wu, Yang Liu, Bo Lang, Lei Huang \\
State Key Laboratory of Software Development Environment, Beihang University, P.R.China\\
\texttt{\small\{wubo,blonster,langbo,huanglei\}@nlsde.buaa.edu.cn} \\
}
\date{}
\maketitle
\begin{abstract}
 Convolutional neural networks (CNNs) can be applied to graph similarity matching, in which case they are called graph CNNs. Graph CNNs are attracting increasing attention due to their effectiveness and efficiency. However, the existing convolution approaches focus only on regular data forms and require the transfer of the graph or key node neighborhoods of the graph into the same fixed form. During this transfer process, structural information of the graph can be lost, and some redundant information can be incorporated. To overcome this problem, we propose the disordered graph convolutional neural network (DGCNN) based on the mixed Gaussian model, which extends the CNN by adding a preprocessing layer called the disordered graph convolutional layer (DGCL). The DGCL uses a mixed Gaussian function to realize the mapping between the convolution kernel and the nodes in the neighborhood of the graph. The output of the DGCL is the input of the CNN. We further implement a backward-propagation optimization process of the convolutional layer by which we incorporate the feature-learning model of the irregular node neighborhood structure into the network. Thereafter, the optimization of the convolution kernel becomes part of the neural network learning process. The DGCNN can accept arbitrary scaled and disordered neighborhood graph structures as the receptive fields of CNNs, which reduces information loss during graph transformation. Finally, we perform experiments on multiple standard graph datasets. The results show that the proposed method outperforms the state-of-the-art methods in graph classification and retrieval.
\end{abstract}

\section{Introduction}

A graph structure is a rich representational form that can describe complex structural data in the real world, such as images, biomedical data, and social networks. Many studies represent an image as an attribute graph and transform the image retrieval problem into an attribute graph search problem. The technique of chemical analysis graph searching, for example, can facilitate the study of properties of newly synthesized chemicals by referring to a database of existing chemicals with known properties. Therefore, it is important to study graph feature learning and searching.

In recent years, deep learning has been applied to many areas and has been shown to significantly outperform traditional methods. Among the available techniques, convolutional neural networks (CNNs) are widely used in image classification, semantic segmentation and object recognition. CNNs can learn the local structure and features of data. Because data such as images, video and sound have the same fixed-sized neighborhoods, convolution, pooling and other operations are well defined in the mathematical sense. For example, in an image, each pixel has eight neighboring nodes. However, traditional CNNs cannot be applied directly to graph data,  whose neighborhoods are irregular.

To apply CNNs to graph-structured data, multiple methods have been proposed~\cite{Bruna2013Spectral_5,Defferrard2016Convolutional_12,Komorowski2012Fluctuations_15,Lin2011Regularized,Simonyan2014Very,RenHG015,Verma2017Dynamic}. These methods can be divided into two categories: spatial-based methods and spectral-based methods. Spatial-based approaches use the neighborhood information from the graph data  space in convolution operations. The main strategy of these methods is to convert the convolution of the graph data into an inner product of the neighborhood information in the graph data space. However, it is challenging to find a convolution operation that is translation-invariant for irregular data. The spectral-based methods typically use the Laplacian to transform the graph data and then use the eigenvector as the convolution operator. The purpose of this transformation is to approximate the convolution operation of the graph data as a convolution operation of the regularized data. In recent work, researchers have attempted to design a graph-CNN architecture by employing a graph-labeling procedure for the construction of a receptive field. Mathias Niepert ~\cite{Niepert2016Learning_10} proposed a framework for learning CNNs for graphs. To a certain extent, the methods mentioned above solve the problem of applying a CNN to graph data. However, all of these methods require the graphs to be transformed into the same neighborhoods with the same ordering . This process is called graph regularization and involves the conversion of graph data into a data format that can be processed by standard CNNs.

The local receptive field of a graph is similar to the fixed-size neighborhood of an image. However, the numbers of neighboring nodes of each node in the graph are not fixed. The standard practice is to regularize the neighborhood of the node: First, a threshold value is fixed. If the neighborhood size is less than the threshold, the neighborhood will be filled with zeros, which is equivalent to adding invalid information. When the neighborhood size is greater than the threshold, we interpret the threshold size of the node as a neighborhood , which results in the loss of some of the effective neighborhood information. However, this approach cannot reflect the neighborhood information of real nodes. This type of model can support continuous labelling with graph data but requires the graph or node neighborhoods of the graph to be transformed into fixed-sized representations to meet the processing requirements of the CNN. Therefore, this method can result in the loss of important information due to the padding and interception operations. Thus, one of the challenges in improving the effect of applying CNNs to graph data is that the neural network model can perform convolution operations directly on irregular node neighborhoods and can perform parametric learning.

To address these limitations, we present a graph convolutional neural network (g-CNN) model that can perform feature learning on graph data directly. We use a continuous mapping function (which is based on a mixed Gaussian process) between the irregular local neighborhood and the convolution weight to transform the discrete parameter learning problem into a parameter sampling problem of a continuous function. Therefore, parameter sampling becomes a function of the features in the preceding layer of the network rather than being based on manually defined parameters on the graph, as in previous studies.

The main innovation of our model is that it does not need to convert the graph or its node neighborhoods into fixed structures; instead, the model learns the irregular structural data directly and can optimize the graph convolution kernel through the neural network. Thus, the model is called the disordered graph convolutional neural network (DGCNN). We conduct experiments on multiple standard graph datasets, and the experimental results show that the proposed method outperforms the existing g-CNN methods and other types of methods in graph classification and retrieval.

The remainder of this paper is organized as follows. In Section 2, we introduce the relevant work on g-CNNs and graph kernels (g-kernels). In Section 3, we introduce the model structure. In Section 4, we introduce the DGCL. In Section 5, we describe the experiment and present the results of our method and the comparison methods. Finally, in Section 6, we discuss the results and present our conclusions , respectively.

\section{Related work}
Current graph processing methods can be divided into two categories: traditional kernel approaches and g-CNNs. The g-CNN approaches often apply standard CNNs to graph data feature learning, while traditional kernel approaches typically use non-linear projection to transform sample graph data into a higher-dimensional feature space, where analysis and processing are performed.

\subsection{G-kernel approaches}
G-kernel approaches project a graph into a feature vector space; the similarity of the two graphs is their scalar product in the space. A g-kernel often defines the similarity function for two graphs. Multiple g-kernels have been proposed, such as the random-walk (RW) kernel, the shortest-path (SP) kernel and the sub-tree kernel.

Gartnerj et al.~\cite{G2003On} proposed an RW kernel function based on computing the RW kernel functions of common steps for two graphs and proved that this function is a positive-definite function. However, the RW g-kernel function has two disadvantages. First, for both of the g-kernels, the comparison of RW paths is of enormous computational complexity. Second, an RW path often contains multiple repeated points and edges, which influences the computational efficiency of RW g-kernel functions. Weisfeiler~\cite{WeisfeilerReduction_26}  proposed the WL sub-tree g-kernel, which is based on the one-dimensional WL isomorphism algorithm. This algorithm searches for the sub-tree structure that is shared by two graphs. However, the WL kernel supports only discrete features, and the memory consumed by the WL kernel is proportional to the number of training samples. The SP kernel (Borgwardt and Kriegel 2006) calculates the similarity by comparing every pair of edges in SP graphs. Shervashidze~\cite{Shervashidze2009Efficient_4} proposed a graphlet count kernel (GK) function based on the sub-graph structure. A graphlet is a small-sized sub-graph that often contains 3 to 5 nodes. Due to the lack of an effective approach for node labelling, this GK function is not applicable to datasets that are focused on node labels.

\subsection{Graph convolutional neural networks}
CNNs are applied to graph data in two broad categories of research: spectral filtering methods and local filtering methods. In the field of spectral filtering methods, Henaff et al. ~\cite{HenaffBL15} used feature vectors of graph Laplacians to perform convolution and used a weighted distance to construct the similarity matrix. Defferrard et al.~\cite{Defferrard2016Convolutional_12} proposed a network model based on ChebNet, which is a spectrally defined method with space attributes. In this model, Chebyshev polynomials of the Laplacian are used to learn k-hop neighborhoods of graph data, thereby incorporating spatial information into neighborhoods. Kipf and Welling~\cite{Kipf2016Semi_14} derived a semi-supervised g-CNN approach by simplifying and extending the ChebNet-based model. All of these approaches require a fixed graph data structure. In the field of local filtering methods, Atwood and Towsley developed a diffusion convolutional neural network (DCNN) that performs RWs in graph data to select the neighborhood structure in the space as the input for the CNN. However, the DCNN is of complexity $O(N^2)$, which restricts the extendibility of this approach. Bruna et al.~\cite{Bruna2013Spectral_5} proposed a multi-scale cluster-based g-CNN model in which convolution defines the weight of each non-shared attribute of each cluster. Duvenaud et al.~\cite{Defferrard2016Convolutional_12} developed a local space filter that can be applied to any node and its neighborhood. Mathias Niepert~\cite{Niepert2016Learning_10} proposed a method that can obtain the local receptive field of graph data and apply it to a CNN, which includes three steps: 1) select a node; 2) construct the fixed-size neighborhood of this node to form a fixed-size sub-graph; and 3) regulate the neighborhood sub-graph. It is possible to obtain a one-dimensional data unit that can be processed by a standard CNN using these three steps. However, both spectrally and spatially defined methods need to transform graph data into data structures with fixed scale, and feature information loss during the transformation process is unavoidable.

Unlike previous work, the g-CNN model we propose, namely, DGCNN, is specially designed for the disordered features of node neighborhoods and can perform convolution from irregular neighborhood structures while achieving the back-propagation of graph convolution without transforming the graph data structure into a fixed regular structure. After parameter sampling based on the Gaussian mixture model (GMM), the DGCNN can perform convolution operations on irregular and disorder neighborhood structures.

\section{Model structure and preprocessing}

The key step for the application of CNNs to normalize grid data is to use a window of size $k*k$ to capture the local neighborhood of the image and share the corresponding convolution kernel parameters in the window. Because of the randomness of the neighborhoods of graph nodes, the traditional window translation method is not applicable to graph data, and a g-CNN model is proposed for accommodating random node neighborhoods.

\begin{figure}[!htbp]
\centering
\subfigure[weights of the image]{\label{weights}
\begin{minipage}[c]{0.5\textwidth}
\centering
\includegraphics[width=4cm,height=4cm]{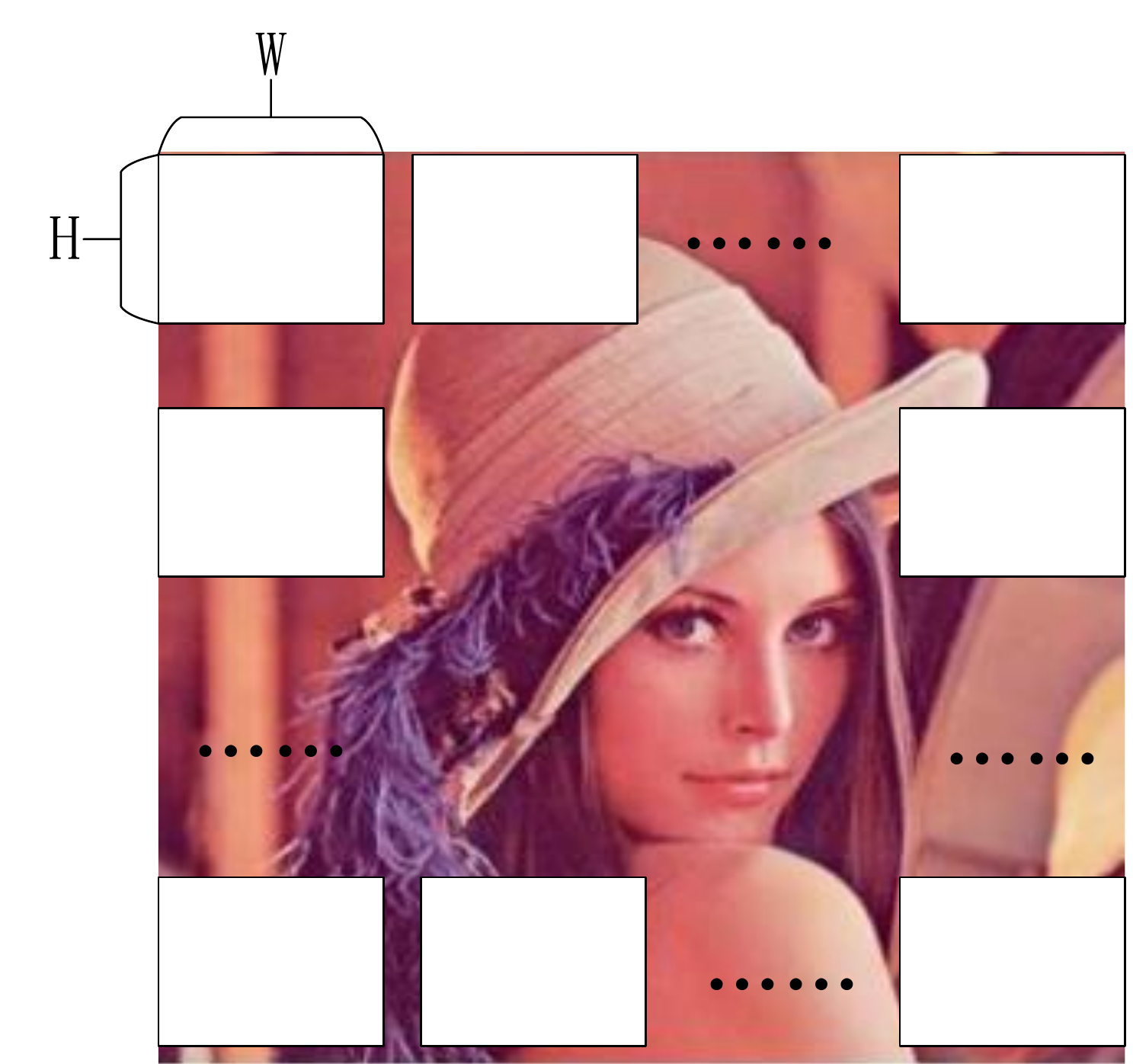}
\end{minipage}%
}%
\subfigure[node neighborhood in the graph data]{\label{neighborhood}
\begin{minipage}[c]{0.5\textwidth}
\centering
\includegraphics[width=4cm,height=4cm]{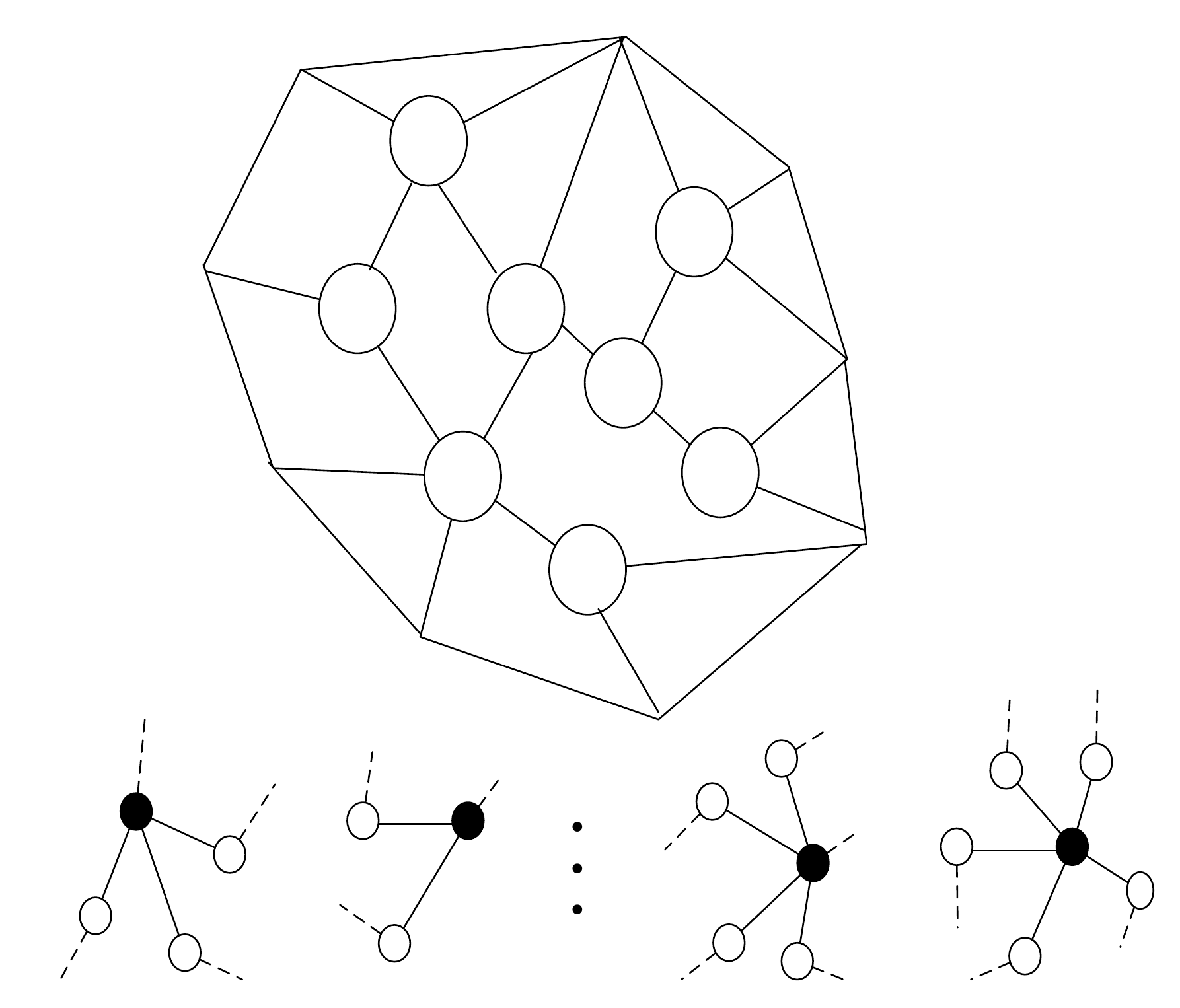}
\end{minipage}
}
\caption{Receptive fields of an image and the corresponding graph}\label{fig:imagetograph}
\end{figure}

When processing an image in the framework of the standard CNN model, the local receptive field is used to implement the convolution operation on the data according to a step movement and obtain the local features of the graph, as shown in Fig.~\ref{weights}, where the convolution window size is fixed to $W*H$. Due to the normalization of the pixel position of the image, the local receptive field can be moved from left to right and from top to bottom to obtain the local information of the image. As shown in Fig.~\ref{neighborhood}, the neighborhoods of the different nodes correspond to various receptive fields of the convolution processes. The node neighborhoods of the graph have no fixed scale or order; thus, the convolution cannot be implemented directly on the graph using the fixed-sized and ordered convolution kernels.

\begin{figure}
\vspace{-0.1in}
\centering
\includegraphics[width=5.5cm,height=7cm]{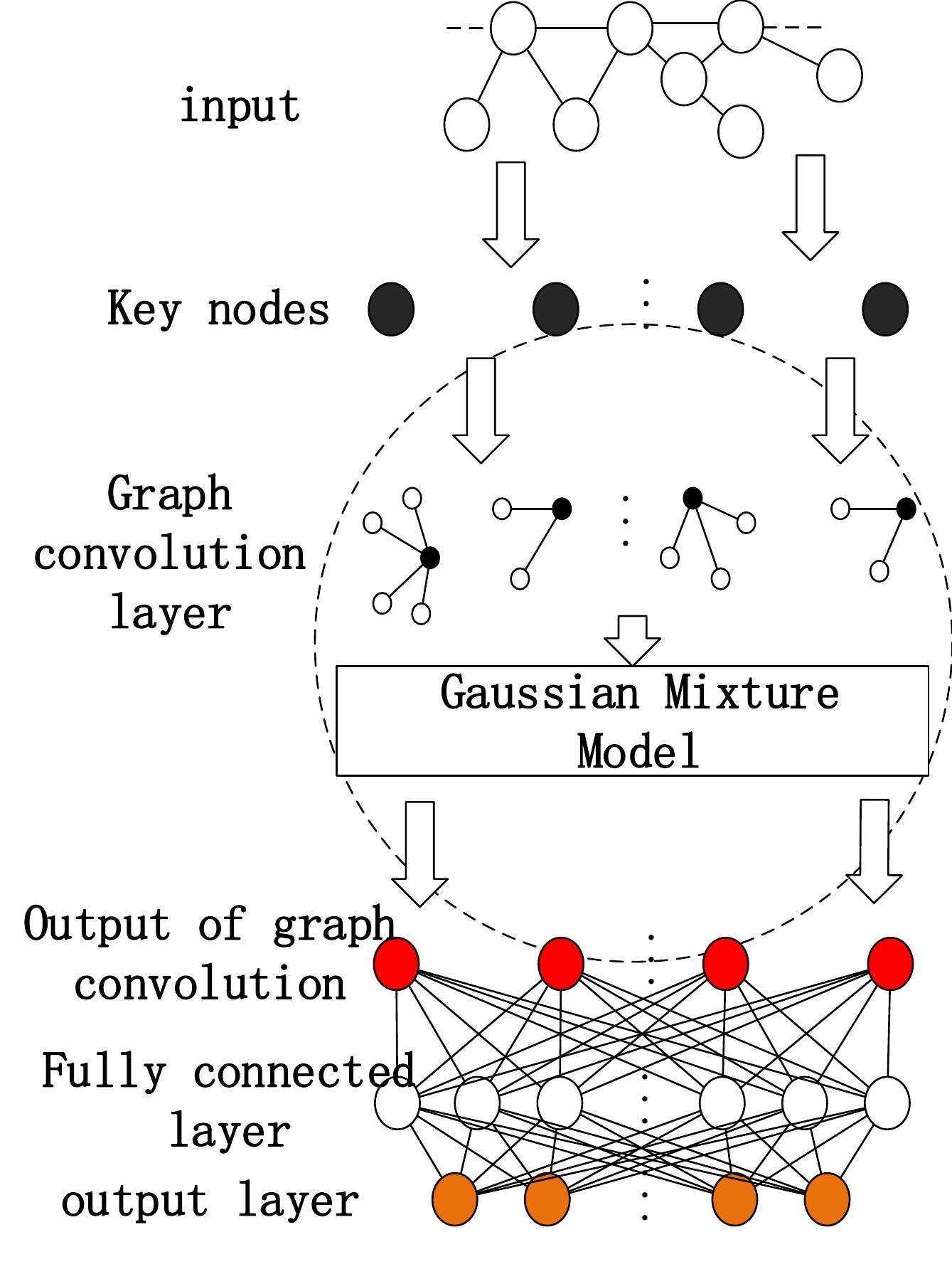}
\vspace{-0.1in}
\caption{Illustration of the proposed architecture}\label{fig:model_gcnn}
\vspace{-0.2in}
\end{figure}

To overcome the above problems, we propose a disordered g-CNN model that can be applied to arbitrary graph data. The network in this model can learn the parameter mapping between the random node neighborhoods and the convolution weights of the graph. As shown in Fig.~\ref{fig:model_gcnn}, the graph data are first transformed into a receptive field that can be processed by the CNNs. Then, a convolution operation is performed over the kernel parameter matrix and the receptive field. Finally, the g-CNN takes the output of the convolutional layer (CL) as the input data of the standard overall connection layer. Our model contains the following parts:

\paragraph{(1)Key Node Selection:} The selection operation is implemented on the graph data to obtain a fixed number of key nodes. To ensure that the number of neighborhoods of the nodes in each graph is consistent, the same number of key nodes is sampled for each graph.

\paragraph{(2)Neighborhood Assembly:} The nodes of the $k-neighborhood$ are the candidates for the receptive field. Note that this time, the receptive field is disordered.

\paragraph{(3)Parameter Sampling:} The corresponding convolution kernel parameters are sampled based on the mixed Gaussian model according to the information of the nodes in each neighborhood to implement the convolution operation for each neighboring graph with its corresponding convolution kernel parameters.

\paragraph{(4)Feature Learning:} By combining the DGCL with the standard CNN and the output layers, the g-CNNs can be built and can directly learn the neighborhood of any random node.

\begin{figure}
\vspace{-0.1in}
\centering
\includegraphics[width=8.5cm]{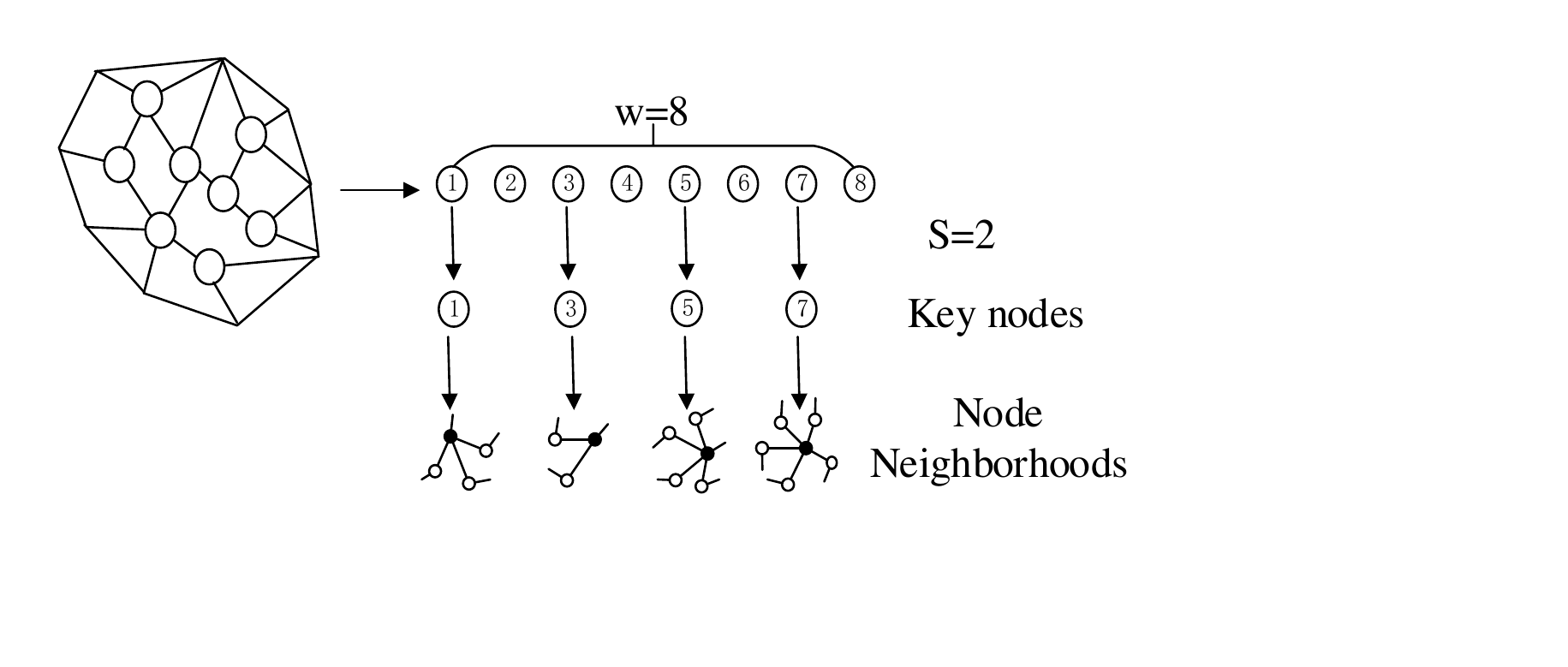}
\vspace{-0.1in}
\caption{Preprocessing procedure of graph data}\label{fig:Preprocessing}
\vspace{-0.2in}
\end{figure}

The preprocessing procedure was implemented on each input graph data, as shown in Fig.~\ref{fig:model_gcnn}, which includes node sampling and node neighborhood construction:

\paragraph{(1)Sequence sampling of key nodes:} To sort all the nodes in the graph, a method proposed elsewhere~\cite{Niepert2016Learning_10} was adopted, and a graph labeling function was introduced in which the set of nodes in the graph are mapped to an ordered node sequence according to the centripetal parameters  (e.g., the node degree or centrad). From the sequence, w nodes are alternately selected according to a certain interval $s$ to form the ultimate node sequences. The nodes in the graph are sorted first, as shown in Fig.~\ref{fig:Preprocessing}, and then four nodes are alternately selected as the key node sequence according to the interval s = 2.

\paragraph{(2)Node neighborhood construction:} As shown in Fig.~\ref{fig:Preprocessing}, for each node in the node sequence that was obtained in the previous step, breadth-first searching is used to find the neighboring nodes, which form the neighborhood set of the original key nodes. The node in each node neighborhood should contain the attribute (such as the weight of the edge or the similarity) between the node and the key node and the attributes of the node (such as the node category).

After the two steps of input graph data preprocessing, the input data are transformed into a random neighborhood set with a fixed size, which is similar to the local receptive field set of the graph.

\begin{figure*}[!htbp]
\centering
\subfigure[illustration of a standard CNN]{\label{fig:CNNs}
\begin{minipage}[c]{0.5\textwidth}
\centering
\includegraphics[width=6cm,height = 4cm]{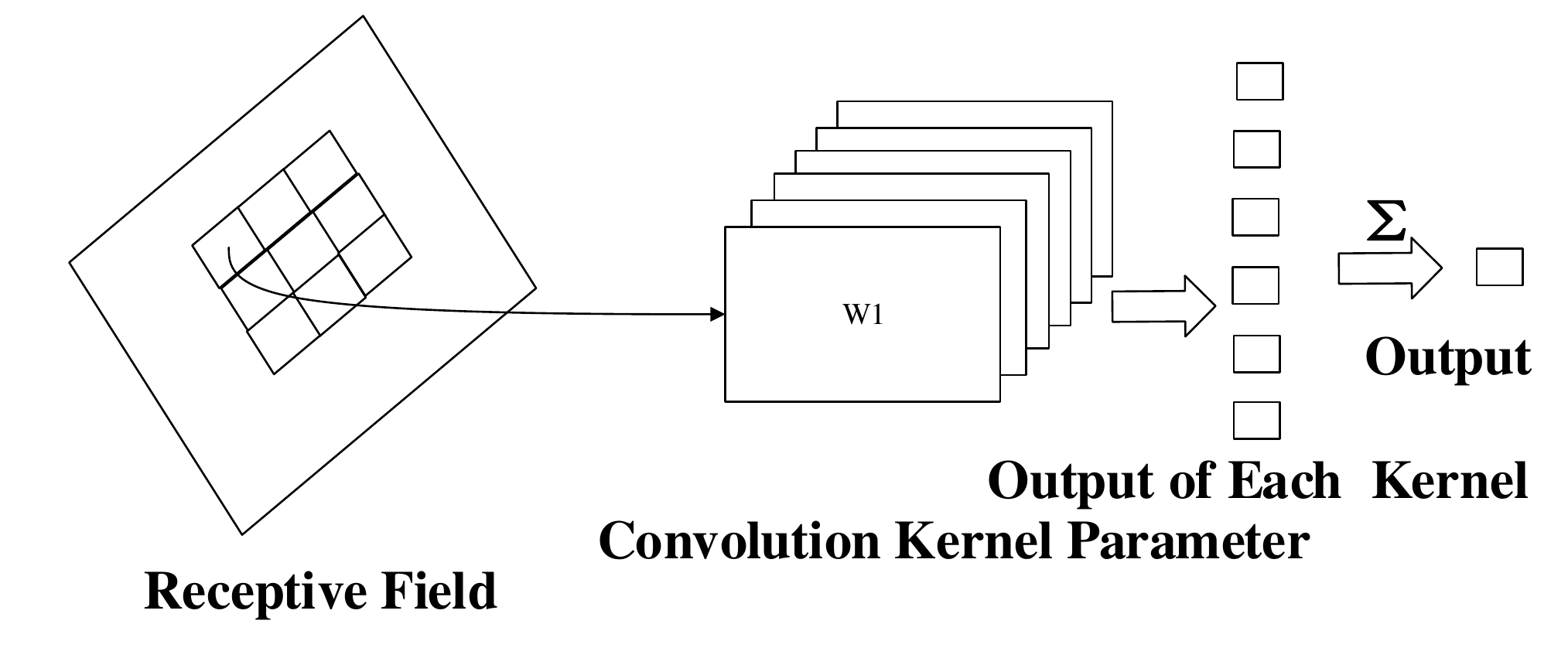}
\end{minipage}%
}%
\subfigure[illustration of a DGCNN based on a GMM]{\label{fig:DGCNNs}
\begin{minipage}[c]{0.5\textwidth}
\centering
\includegraphics[width=6cm,height = 4cm]{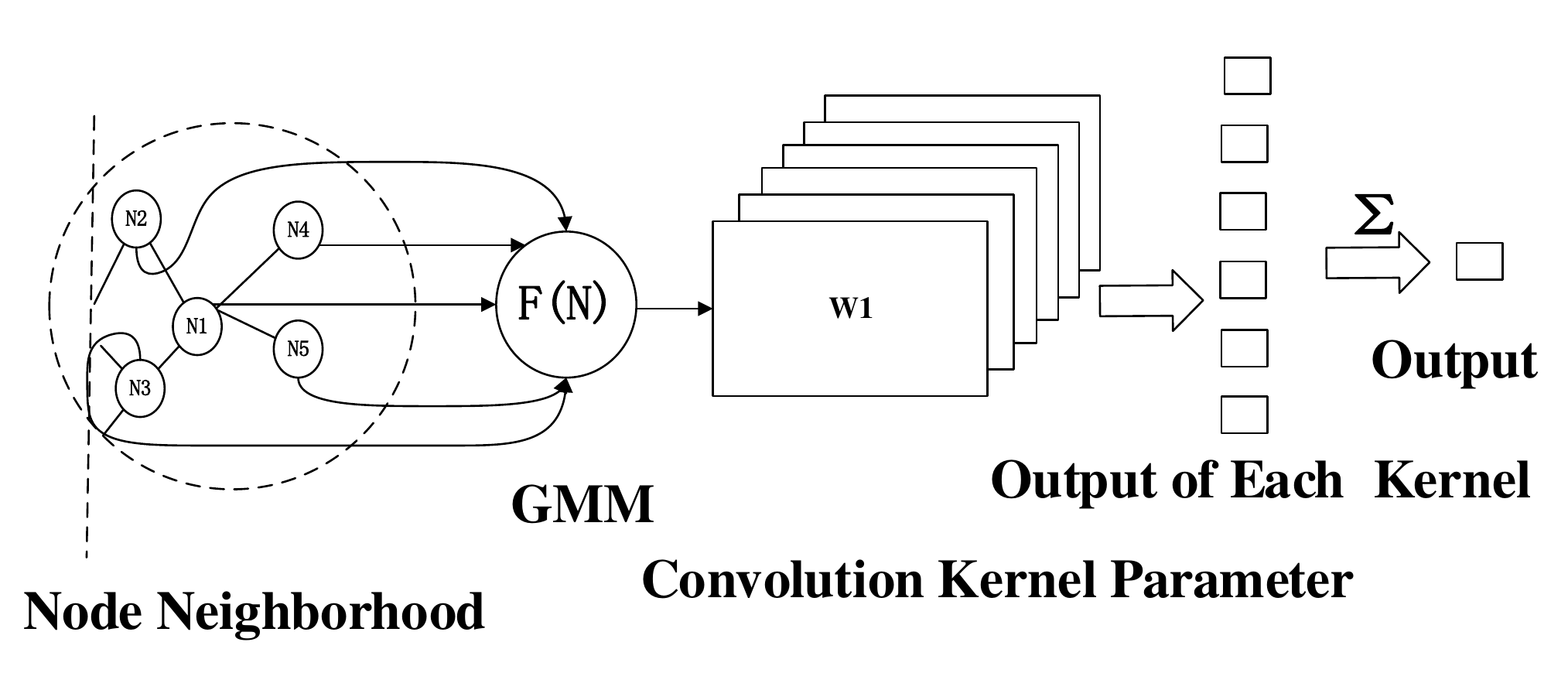}
\end{minipage}
}
\caption{Convolutional Process Comparison between the Standard Image and a Node Neighborhood of the Graph Data}\label{fig:DGCNNsandCNN}
\end{figure*}

\section{DGCL and its learning process}
In the DGCNN, a disordered CL that can receive and process any graph data is designed. A disordered CL is a CL that can perform convolution operations on irregular and disordered node neighborhoods while achieving the back-propagation of the CNN.

\subsection{Disordered graph convolutional layer}
For a DGCL, the input is the node neighborhood structure that was obtained after graph preprocessing. With the GMM and an activation function, the input graph is transformed into the output of the CNNs.

Fig.~\ref{fig:CNNs} shows a standard CNN with a receptive field on an image. The receptive field is of fixed size and ordered. For graph data, each node neighborhood is of variable size and disordered, as shown in Fig. \ref{fig:DGCNNs}. For example, the node neighborhood in Fig.~\ref{fig:DGCNNs} includes 5 nodes. Among nodes N1, N2, N3, N4 and N5, N1 is the key node, and the others are its neigh boring nodes. It is necessary to obtain the convolutional kernels for these five nodes before performing the convolution operation. An existing solution is to define a fixed-sized convolutional kernel parameter, which requires a regularization process that may lead to information loss. However, we sample the convolutional parameters of each neighborhood node on the possibility distribution for the similarity of the neighborhood nodes and the key node, and the number of convolutional parameters is the same as the number of key node neighborhoods. According to the central limit theorem~\cite{Cs2002Almost}, it is reasonable to assume that the probability distribution of the parameters is defined by the GMM, which can approximate any probability distribution. Thus, for such a node neighborhood, we sample the convolutional parameters on the sheaf of a Gaussian function GMM~$(\theta)$ based on the GMM for each node neighborhood. The output value of the convolution operation for such a neighborhood and the sampled kernel parameter is:

\begin{equation}
F(N) =(GMM(\theta),X) = \sum_{k=1}^5 (\sum_{i=1}^n w_iG(\theta_k,\mu_i,\sigma_i),X_k)
\end{equation}

where $N$ denotes the neighborhood map of a key node, $X$ is the attribute value of a node according to this map, parameter $X_k$ is the attribute of the $k-th$ node, $\theta$ is the correlation between the $k-th$ node and the key node, $m$ is the number of Gaussian components, $w_i$ is the weight of each Gaussian component, $\mu_i$ and $\sigma_i$ are the mean value and variance of each Gaussian component, respectively, and $G(\cdot)$ is the Gaussian function.

After sampling the convolutional parameters for all key nodes, convolution operations are performed on the neighborhood to finish the graph data convolution in the graph CL. The convolutional processing of the graph data represents the forward propagation of the neural network, as shown in Fig.~\ref{fig:DGCNNs}.

\begin{figure}
\vspace{-0.1in}
\centering
\includegraphics[width=6cm,height=4cm]{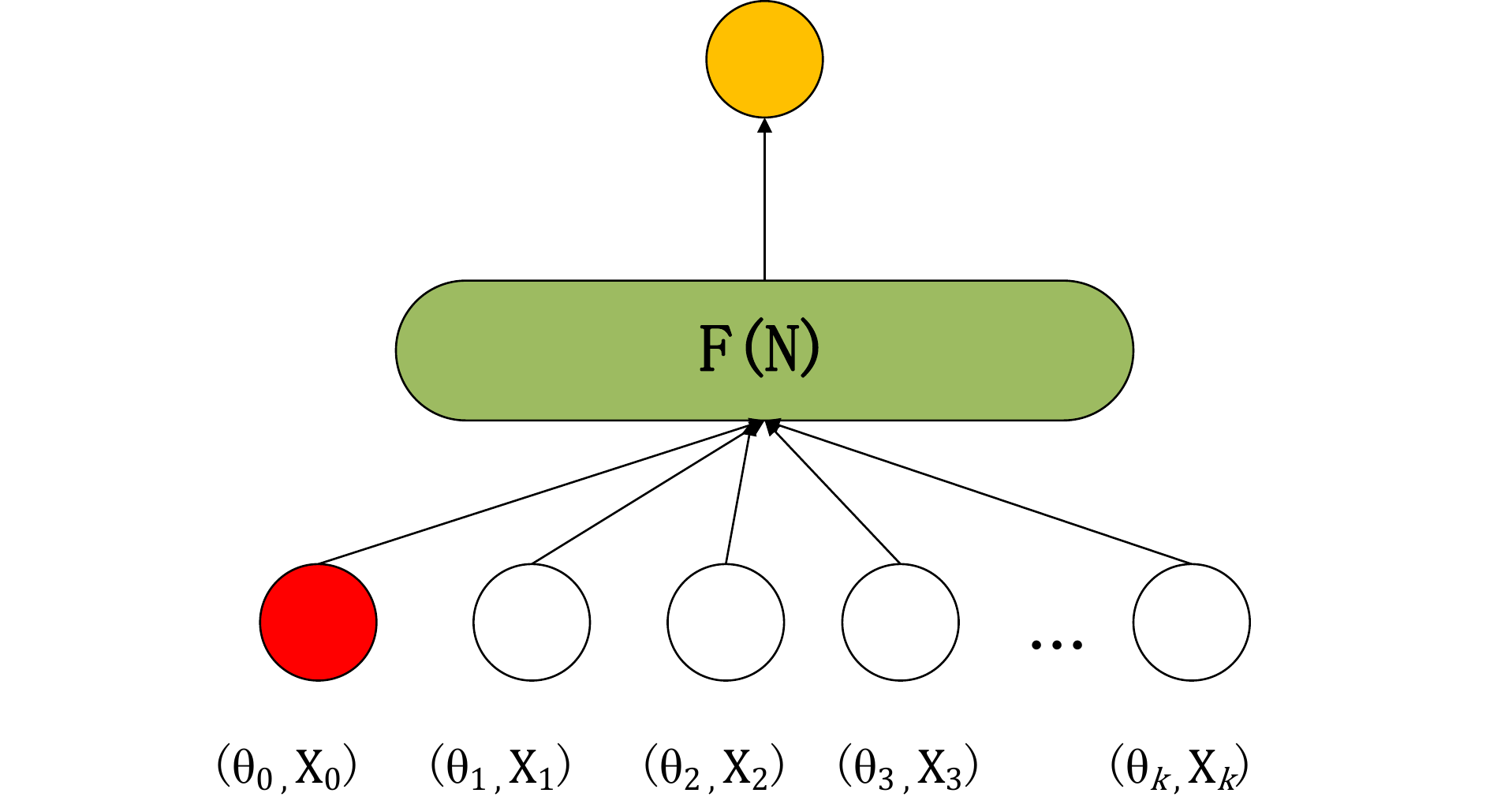}
\vspace{-0.1in}
\caption{Convolutional Unit of A Node Neighborhood}\label{fig:CUAN}
\vspace{-0.2in}
\end{figure}

In Fig.~\ref{fig:CUAN}, the red node in the input part represents the key node, the white nodes are the neighboring nodes of the key node, the yellow node is the output node, and the green oval in the middle represents $F(N)$. $\theta_0$ is a constant (in this study, $\theta_0=0$), and $\theta_1, \theta_2, \cdots, \theta_k$ represent the attribute values of edges between neighboring nodes and the key node, and $X_0, X_1, X_2,..., X_k$ are the attribute values of the nodes.

As for the j-th receptive field, we can obtain the forward output for which the graph convolutional process is performed on its receptive field.

\begin{equation}
f_j=\sum_{i=0}^I GMM(\theta_i)X_i+b
\end{equation}

where $I$ denotes the size of the neighborhood sub-graph, $X_i$ denotes the attribute value of the i-th node, $GMM (\theta_i)$ denotes mixed Gaussian values of the i-th node, and $b\in R^E$ denotes a vector of bias terms.

\subsection{Graph convolution back-propagation and Gaussian parameter learning}
In the DGCL based on the GMM, the parameters of each component of the GMM must be optimized. The difference between the output value of CNN and the real value is then used for back-propagation to adjust the parameters. The error function for back-propagation is defined in formula (3)~\cite{NIPS2012_4824}:

\begin{equation}
\begin{split}
error(\theta) &= \frac{1}{2} \sum_{a=1}^A \sum_{b=1}^B(t_b^{(a)}(\theta) - f_b^{(a)}(\theta))^2 \\
              &= \frac{1}{2} \sum_{a=1}^A \sum_{b=1}^B(\nabla f_b^{a}(\theta))^2
\end{split}
\end{equation}

where  $t_i^{(a)}$ is the $a$-th dimension of the corresponding label of the $b$-th graph data, $f_b^{(a)}$ is the similarly value of the $a$-th output layer unit in response to the $n$-th input pattern, $B$ is the number of graph data types, and $A$ is the number of graph data.

For the $a$-th graph data, we can immediately compute the gradient:

\begin{equation}
\begin{split}
\nabla f_\theta^{a} &= \frac{\partial f^{a}}{\partial(w_1,\mu_1,\sigma_1,...,w_m,\mu_m,\sigma_m)} \\
&=(\frac{\partial f^{a}}{\partial w_1},\frac{\partial f^{a}}{\partial \mu_1},\frac{\partial f^{a}}{\partial \sigma_1},...,\frac{\partial f^{a}}{\partial w_m},\frac{\partial f^{a}}{\partial \mu_m},\frac{\partial f^{a}}{\partial \sigma_m})
\end{split}
\end{equation}

where $f$ denotes the output of forward-propagation; $w, \mu, \sigma$ are the weight, mean value and variance of each Gaussian component, respectively; and $m$ is the number of Gaussian components. We need to calculate the derivative and parameters of the Gaussian component:

\begin{equation}
\frac{\partial f}{\partial w_i} = \frac{1}{\sqrt{2\pi}\sigma_i}e^{-\frac{(x-\mu_i)^2}{2\sigma_i^2}}
\end{equation}

\begin{equation}
\frac{\partial f}{\partial \sigma_i} = \frac{w_i}{\sqrt{2\pi}\sigma_i}e^{-\frac{(x-\mu_i)^2}{2\sigma_i^2}}[-1+\frac{(x-\mu_i)^2}{\sigma_i^2}]
\end{equation}

\begin{equation}
\frac{\partial f}{\partial \mu_i} = -\frac{{w_i}*(x-\mu)}{\sqrt{2\pi}\sigma_i^3}e^{-\frac{(x-\mu_i)^2}{2\sigma_i^2}}
\end{equation}

\begin{equation}
w_{i+1} = \lambda*\frac{\partial f}{\partial w_i}
\end{equation}

\begin{equation}
\sigma_{i+1} = \lambda*\frac{\partial f}{\partial \sigma_i}
\end{equation}

\begin{equation}
\mu_{i+1} = \lambda*\frac{\partial f}{\partial \mu_i}
\end{equation}

where  $w_{i+1},\mu_{i+1},\sigma_{i+1}$ are the parameters that are obtained after updating $w_i,\mu_{i},\sigma_{i}$, respectively, and $\lambda$ is a learning rate parameter. In practice, there is often a learning rate parameter $\lambda$ for each Gaussian component.

The computation of the gradient of bias $b$ is the same as that for the traditional CL and is explained here.

\subsection{Number of parameters and computational complexity}
Each weight matrix $W$ is obtained by sampling the mixed Gaussian model, and the number of weight matrices is equal to the number of neighboring nodes of key nodes. The parameters in our method with respect to a conventional CNN are the Gaussian component weight $w$, mean value $\mu$, and variance $\sigma$ for each vector. Let $N$ be the number of nodes in the graph and $M$ be the number of Gaussian components. The total number of parameters is $3*N*M$. Here, we ignore the bias terms, which contribute few parameters.

The complexity of the DGCNN consists of the forward- and back-propagation complexities. Let $k$ be the number of key nodes in the graph. Let $E$ denote the average number of neighbors of each node. For the forward-propagation process, DGCNN has a worst-case complexity of $O(f(k*M*E))$, and for the back-propagation process, DGCNN has a worst-case complexity of $O(3*f(k *M*E))$. Let $T$ be the number of graphs. The total computational complexity is $O(4*T*f(k*M*E))$.

\section{Experiment}
\subsection{Graph datasets}

We conduct our experiments on the following popular real datasets to compare our approach with state-of-the-art g-kernels and CNNs in terms of retrieval precision:

\begin{itemize}
\item PTC~\cite{Toivonen2003Statistical}: PTC consists of 344 chemical compounds, where the classes indicate carcinogenicity for male and female rats.
\item D\&D~\cite{Wang2002Distinguishing}: D\&D is a data set of 1178 protein structures classified into enzymes and non-enzymes.
\item AIDS~\cite{DBLP:journals/corr/WallachDH15}: The dataset contain 4395 chemical compounds, of which 423 belong to class CA, 1081 to CM, and the remaining compounds to CI .
\item PROTEIN~\cite{Niepert2016Learning_10}: PROTEINS is a graph collection in which nodes are secondary structure elements and edges indicate neighborhoods in the amino-acid sequence or in 3D space. Graphs are classified as enzymes or non-enzymes.
\item COLLAB~\cite{Koren2008Factorization}: The dataset contains 12,000 graphs, each with an average of 400 nodes. The dataset contain users, movies, and the users’ ratings of the movies.
\end{itemize}

\subsection{Experimental configuration}
(1) We compare the DGCNN method that is proposed herein with the following methods by experiments on graph classification and graph search:
\begin{itemize}
\item G-kernel method: the SP kernel~\cite{Borgwardt2006Shortest1}, the RW kernel~\cite{G2003On}, the GK kernel~\cite{Orsini2015Graph}, and the Weisfeiler-Lehman subtree kernel (WL)~\cite{WeisfeilerReduction_26}.
\item g-CNNs: PATCHY-SAN~\cite{Niepert2016Learning_10}, ~LMFGCN~\cite{Duvenaud2015Convolutional}, and SSGCN~\cite{Kipf2016Semi_14}.
\end{itemize}

(2) We test the influence of the number of Gaussian components on the proposed model using 5, 10, 15, 20, 25, 30 and 35 components and find the optimal number, at which the best effect is obtained;

(3) We perform efficiency analysis.

In our experiment, when calculating the Weisfeiler-Lehman fingerprint, the number of iterations is set as $h=10$, the GK parameter is set as 7, and the factor of RW is set as $10^{-6},10^{-5},…,10^{-1}$. For PATCHY-SAN, the fixed receptive field in this paper is $k=5$. The network structure consists of two CLs, which are one-dimensional ($5*1$); one dense hidden layer; and one softmax layer. The CNN uses $3*3$ filters. The network structure of SSGCN is consistent with that of PATCHY-SAN.
To obtain fair comparison results, for the graph classification experiment, the network structure in DGCNN consists of one DGCL, one standard CL, one dense hidden layer and one softmax layer. In the graph search experiment, the network structure is the same as that used for graph classification. In this study, the output of the dense hidden layer is treated as a feature vector of the graph data.

All the experiments are performed under the following configuration: an Intel Xeon X5650, a dual-core CPU running at 2.66 GHz  with 8 GB memory, and a Linux system. The methods that use CNNs are implemented using the torch framework.

\begin{figure}[t]
\hspace{0.6in}
\subfigure[PTC]{\label{fig:fft:a}
\begin{minipage}[c]{0.5\textwidth}
\centering
\includegraphics[width=5cm]{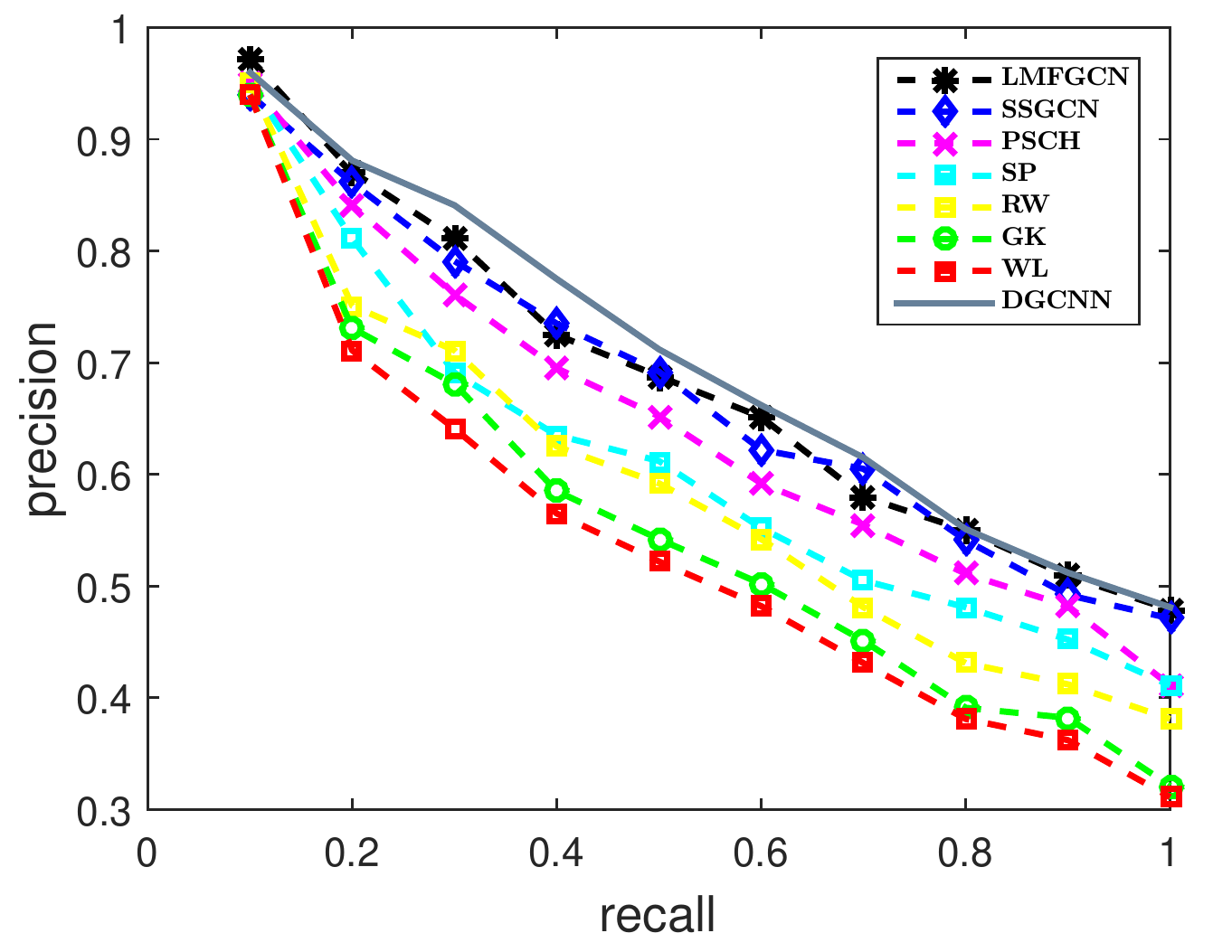}
\end{minipage}%
}%
\hspace{-1in}
\subfigure[PROTEIN]{
\begin{minipage}[c]{0.5\textwidth}
\centering
\includegraphics[width=5cm]{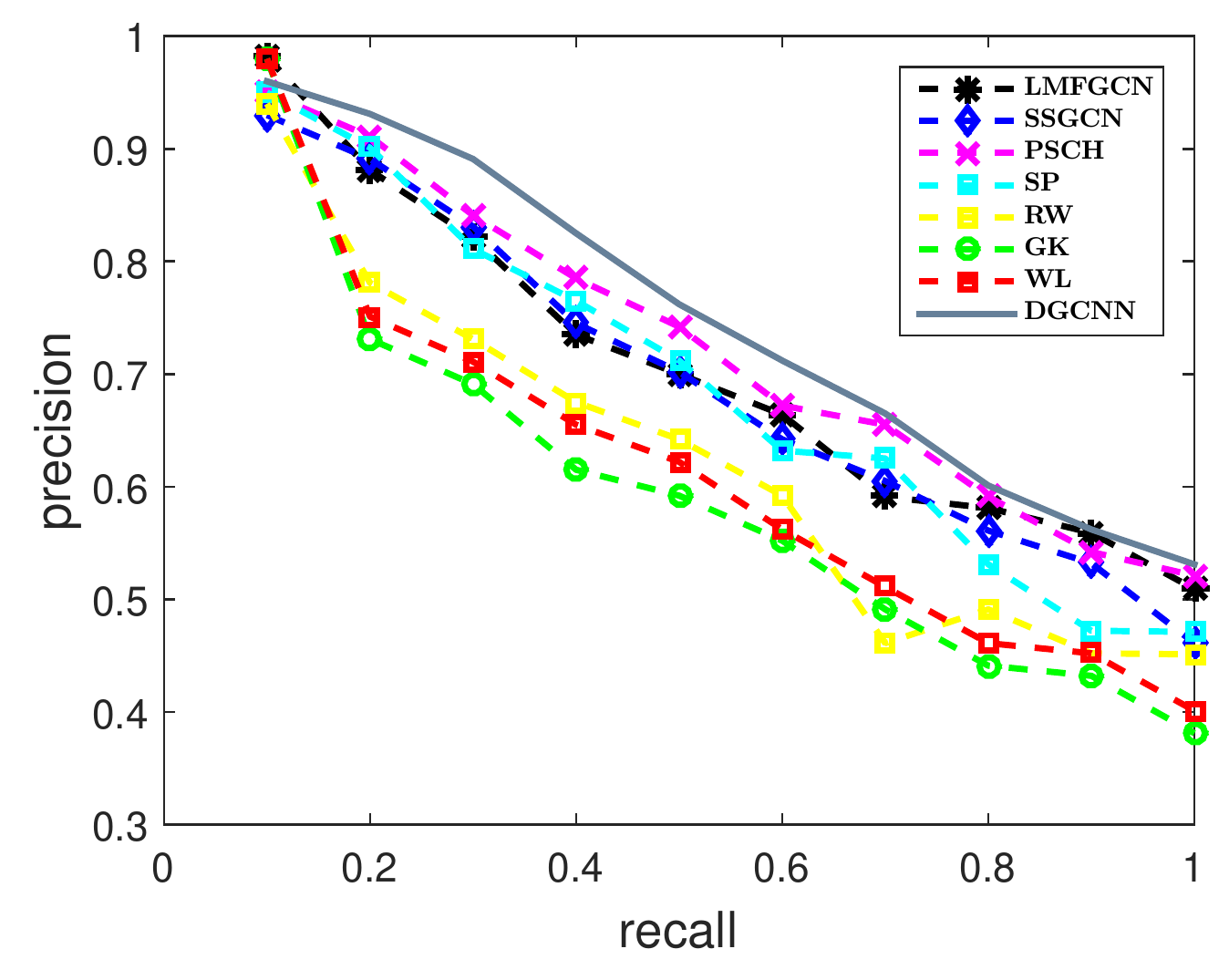}
\end{minipage}
}

\hspace{0.2in}\subfigure[AIDS]{
\begin{minipage}[c]{0.3\textwidth}
\centering
\includegraphics[width=5cm]{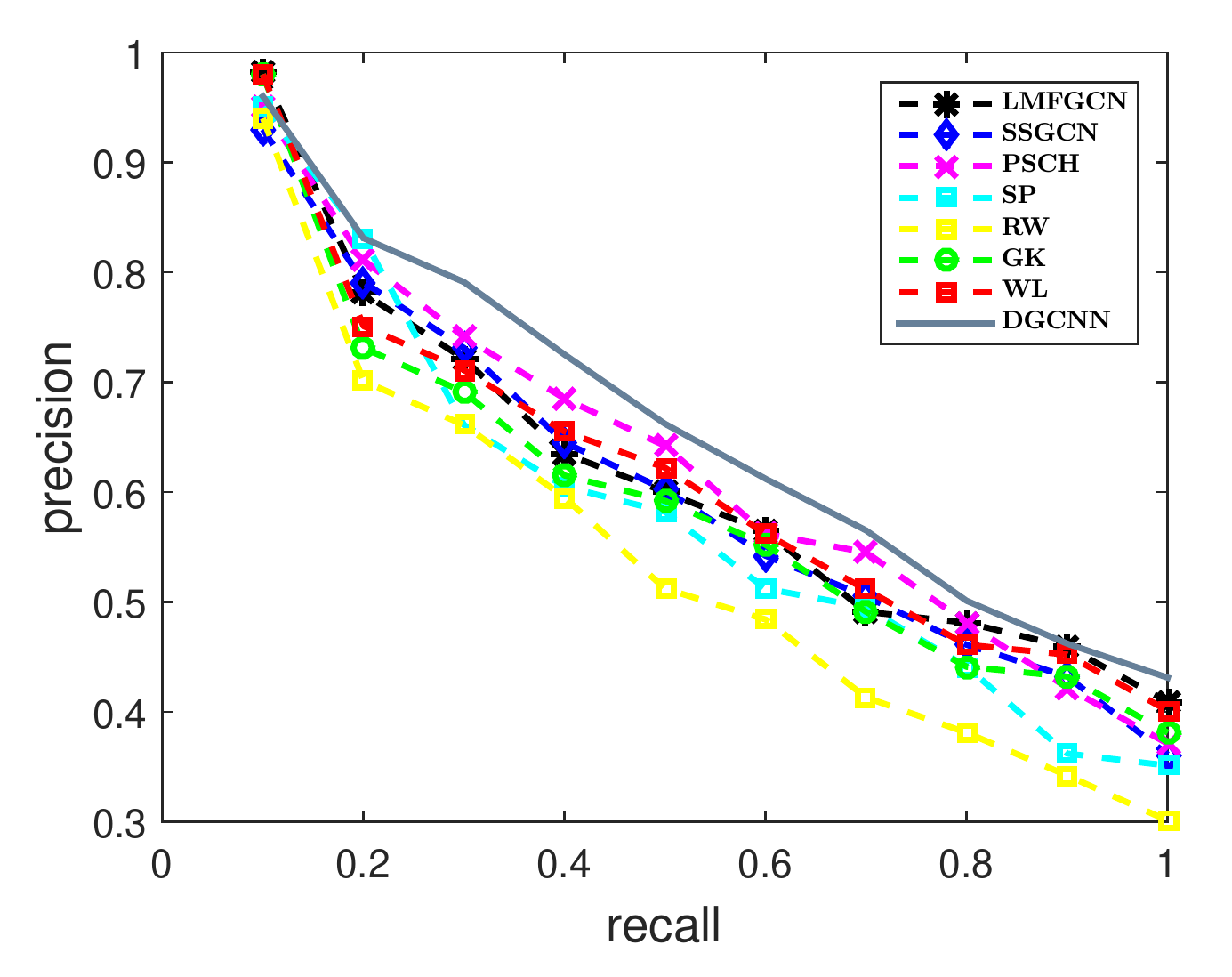}
\end{minipage}
}
\subfigure[D\&D]{
\begin{minipage}[c]{0.3\textwidth}
\centering
\includegraphics[width=5cm]{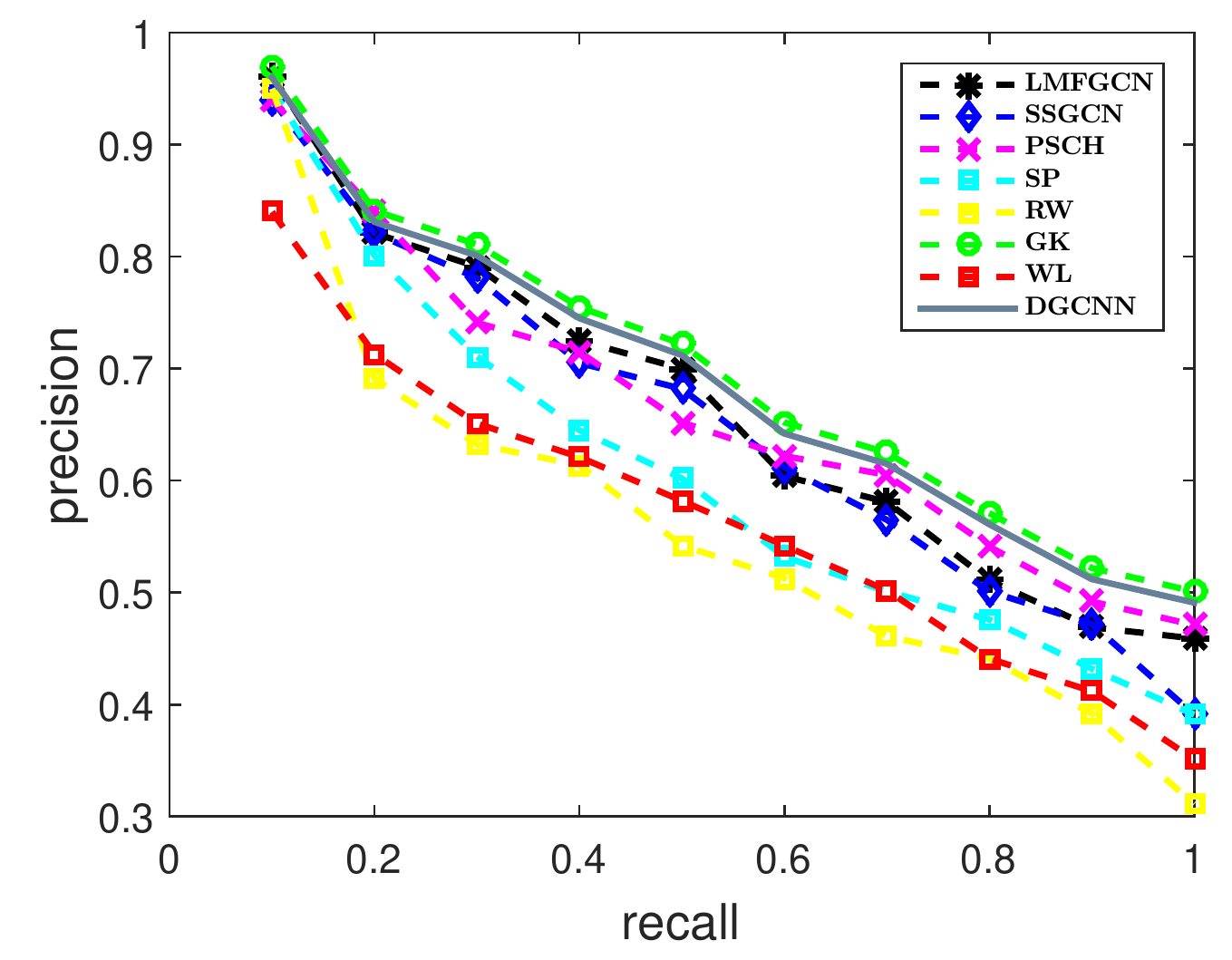}
\end{minipage}
}
\subfigure[COLLAB]{
\begin{minipage}[c]{0.3\textwidth}
\centering
\includegraphics[width=5cm]{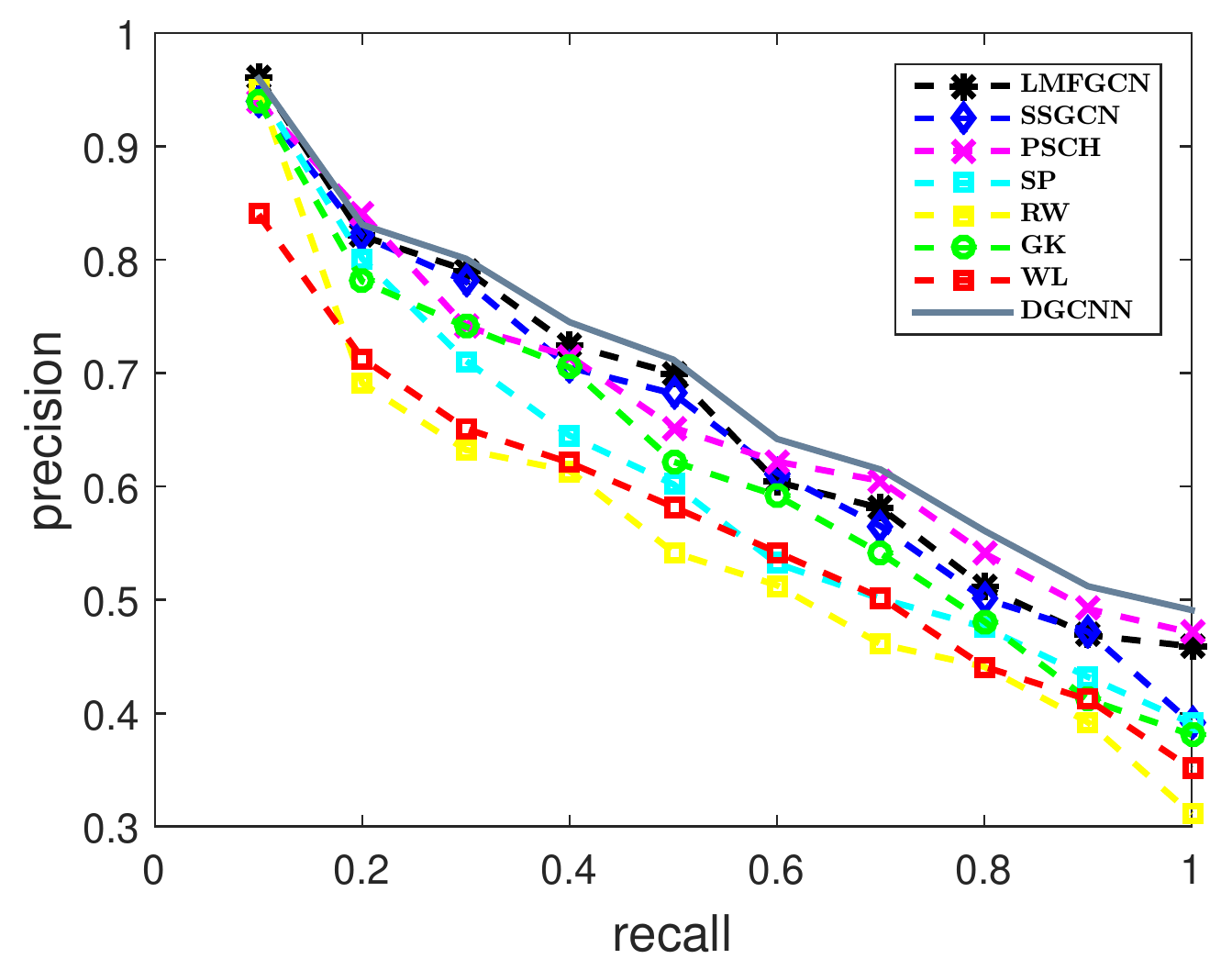}
\end{minipage}
}

\caption{Retrieval precision on five graph datasets for DGCNN, graph kernel methods and recent graph convolution networks}\label{fig:retrieval}
\end{figure}

\subsection{Experimental results}
(1) Comparison of the Graph Classification Results

\begin {table*}[!htbp]
\small
 \caption{\label{tab:test}Comparison of Classification precision between four graph kernel and two graph CNN Methods on Multiple Graph Datasets} \centering
 \vspace{0.2in}
 \begin{tabular}{cclccc}
  \toprule
  Dataset & PTC & AIDS & PROTEIN & D\&D &  COLLAB \\
  \midrule
 number of graphs & 344 & 4395 & 600 & 1178  & 12000 \\
 Max & 109 & 207 & 620 & 5748  & 4123 \\
 Avg & 25.56 & 30.15 & 39.06 & 284.32  & 400 \\\hline
 SP & 58.53$\pm$2.55 & 62.37$\pm$1.13&65.12$\pm$1.01&71.00$\pm$1.11 &61.23$\pm$2.12 \\
 RW & 57.26$\pm$1.30 & 58.37$\pm$2.21&68.11$\pm$2.02&68.10$\pm$0.11 & 65.20$\pm$3.21 \\
 GK & 57.32$\pm$1.13& 60.27$\pm$1.12&61.12$\pm$1.03&\textbf{78.45$\pm$0.26}& 64.45$\pm$3.12 \\
 WL & 56.97$\pm$2.01 & 59.97$\pm$1.01&62.23$\pm$1.03&77.95$\pm$0.70 & 61.25$\pm$1.72 \\\hline
 PSCH~\cite{Niepert2016Learning_10} & 59.43$\pm$3.14 & 60.10$\pm$2.72&72.10$\pm$1.72&74.58$\pm$2.85 & 62.32$\pm$2.45\\
 LMFGCN~\cite{Duvenaud2015Convolutional} & 61.32$\pm$1.21 & 59.21$\pm$2.32 & 67.12$\pm$2.25 & 73.13$\pm$1.13 & 66.13$\pm$2.01\\
 SSGCN~\cite{Kipf2016Semi_14} & 60.21$\pm$2.14 & 55.10$\pm$1.15&65.10$\pm$3.12&75.10$\pm$1.02 & 63.10$\pm$ 2.12 \\\hline
 DGCNN & \textbf{65.43$\pm$3.14} & \textbf{65.10$\pm$1.82}& \textbf{75.10$\pm$2.72} &77.21$\pm$0.85 & \textbf{68.34$\pm$3.13} \\
  \bottomrule
 \end{tabular}
\end{table*}

Table~\ref{tab:test} shows the graph classification accuracy on each dataset for eight methods: SP, RW, GK, WL, PSCH, LMFGCN, and DGCNN. The DGCNN, the method presented here, demonstrated good accuracy on most datasets and attained its best results on the PTC, AIDS, COLLAB and PROTEIN datasets. Unlike traditional g-kernel approaches, the DGCNN exhibits a significant advantage on most datasets. Existing g-CNNs use the standard CNN model and predefine only a one-dimensional convolutional kernel (e.g., $1*5$) in the convolutional feature-learning process without making full use of the space information between the key nodes and their neighborhood information. The method proposed here transforms the discrete learning process into a process of projecting continuous functions and fully learns the relationships between the neighboring nodes and key nodes in the graph data. Therefore, on the most datasets, DGCNN outperforms the existing g-CNN methods and g-kernel methods.

(2) Comparison of the Graph Retrieval Results

The experiment compares the DGCNN and existing methods in terms of graph similarity retrieval performance. We use the output of the dense hidden layer as the feature vector of the graph data. To ensure a fair comparison, for PATCHY-SAN, we also use the output of the dense hidden layer as the feature vector of the graph data. As indicated by the experiment results shown in Fig.~\ref{fig:retrieval}, on the PTC, PROTEIN, COLLAB and AIDS datasets, the DGCNN outperforms the other methods. On the D\&D dataset, the DGCNN method performs similarly to the optimal GK method. The DGCNN outperforms the recently proposed g-CNN methods on most datasets because the process of regularizing node neighborhoods leads to loss of information about the node neighborhoods, whereas our convolutional kernel method, which is based on the GMM, builds a dynamic graph convolutional kernel, which eliminates the local information loss during node neighborhood regularization.

(3) Influence of the Number of Gaussian Components

\begin{figure}
\vspace{-0.1in}
\centering
\includegraphics[width=8cm]{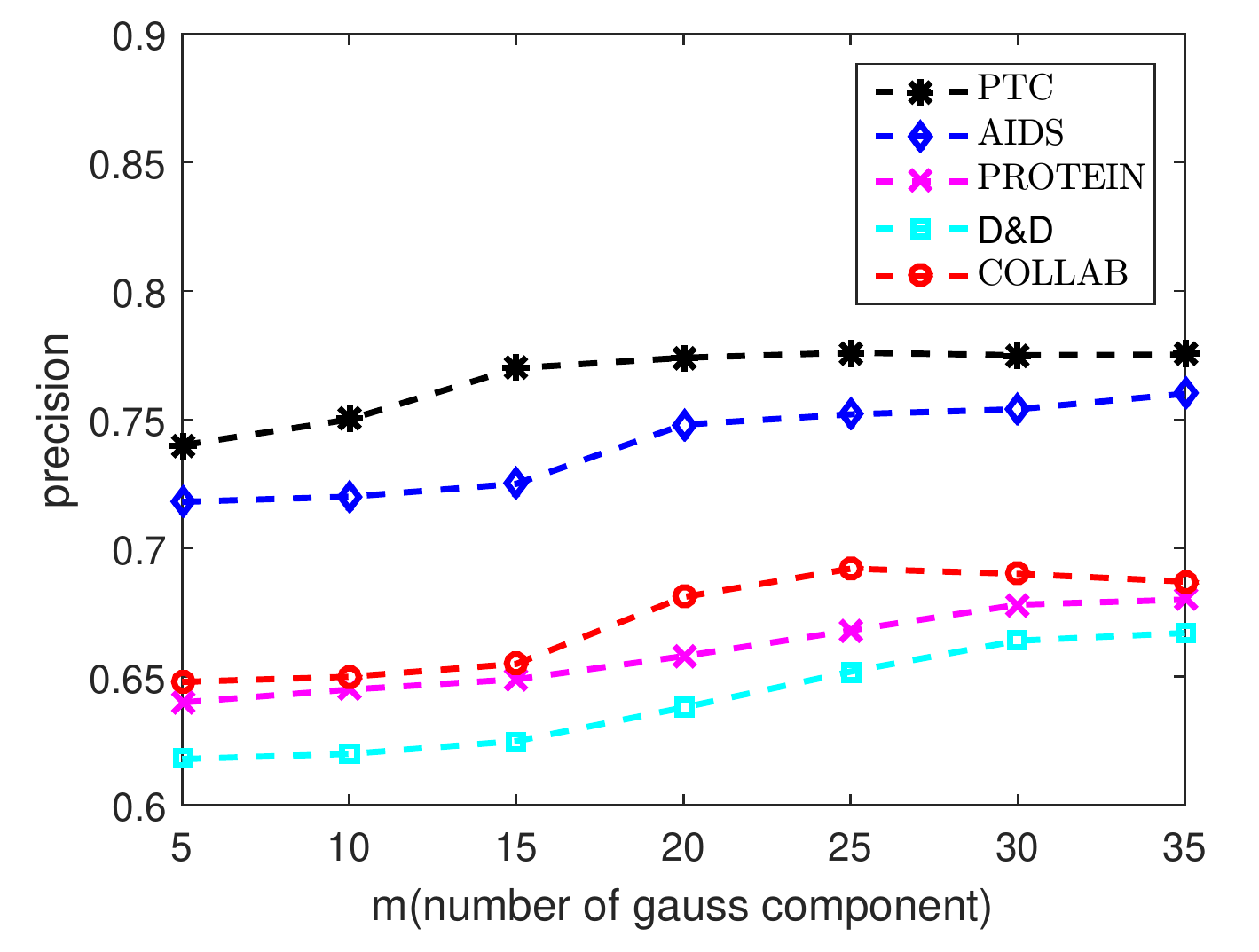}
\vspace{-0.1in}
\caption{Relationship between Number of Gaussian Components and Classification precision for Different Datasets}\label{fig:number}
\end{figure}

The purpose of this experiment is to examine the influence of the number of Gaussian components on the DGCNN. In the experiment, the number of Gaussian components is defined as $m$, which is set as 5, 10, 15, 20, 25, 30 or 35. For each case, we perform the same experiment on multiple graph datasets and calculate the mean result as the final result. The experimental results are shown in Fig.~\ref{fig:number}.

\begin {table*}
\small
 \caption{\label{tab:test_time}Comparison of running time on five graph datasets (in seconds)} \centering
 \vspace{0.2in}
 \begin{tabular}{cclccc}

  \toprule
  Dataset & PTC & AIDS & PROTEIN & D\&D & COLLAB \\
  \midrule
 WL~\cite{WeisfeilerReduction_26} & 30  & 65 & 143 & 609 & 245 \\
 PSCH~\cite{Niepert2016Learning_10} & 6 & 10 & 31 &154 & 235\\
 LMFGCN~\cite{Duvenaud2015Convolutional} & 14 & 23 & 41 & 72 & 62\\
 SSGCN~\cite{Kipf2016Semi_14} & 15 & 20 & 58 & 150 & 200 \\\hline
 DGCNN & 16 & 25 & 57 & 152 & 212 \\
  \bottomrule
 \end{tabular}
\end{table*}
For dataset PTC, the classification precision increases significantly from $m = 5$ to $m = 15$ and is stable when $m$ is larger than 15. For ADIS, the classification precision increases significantly from $m = 5$ to $m = 20$ and is relatively stable when m is larger than 20. Similarly, for the PROTRIN and D\&D datasets, the classification precision increases significantly from $m = 5$ to $m = 15$ and from $m = 5$ to $m = 25$, respectively, and is stable when m is larger. For the dataset with many graphs, namely COLLAB, more parameters are needed to make the model fit the data. We found that $m = 25$ gives the best results on this graph dataset.

These experiments indicate that the optimal numbers of Gaussian components for different datasets are different. Increasing the number above the optimal number does not increase the precision.

In the next experiments, we use the optimal number of Gaussian components with the best classification result for each dataset.

(4) Efficiency analysis

We assess the efficiency of the DGCNN by applying it to graph datasets. For a given graph, we compute the end-to-end running time. The results in Table~\ref{tab:test_time} show that the running time of the DGCNN is less than that of WL. In addition, the average values for other graph convolutional approaches on different datasets are 11.7 s (for PTC), 18 s (for AIDS), 44 s (for PROTEIN), 126 s (for D\&D), and 165 s (for COLLAB). The results of the DGCNN are somewhat slower (the gap is 4.3 s for PTC, 7 s for AIDS, 13 s for PROTEIN, 26 s for D\&D, and 47 s for COLLAB) than the other graph convolutional approaches, probably because generating the receptive fields takes less time than sampling the kernel parameters on the GMM. The recent graph convolution approaches need to generate the receptive fields as the input of the CNN. In the CL of our approach, additional computation is required to sample the parameters on the GMM. However, the total running time of these additional computations is not significantly different and is of the same order of magnitude.

\section{Discussion and conclusions}

We have presented a DGCNN based on a GMM that is applicable to graph similarity matching. The main innovation of this model is that by sampling the corresponding convolutional kernel parameters from a mixed Gaussian distribution, the dynamic convolutional kernel is adapted to the size and order of the node neighborhood. Therefore, our model supports different scales of convolutional receptive fields, thereby avoiding the loss of graph information during the regularization of node neighborhoods. The key aspect of our model is the GMM-based DGCL, which performs convolutional learning on local node neighborhoods of any graph while achieving the back-propagation of graph convolution.

Graph convolutional parameters are combined into a neural network optimization process and optimized to a large degree. Finally, we perform graph classification and search experiments on standard graph datasets such as PTC, PROTEIN, COLLAB and AIDS. These experiments indicate that the DGCNN outperforms existing g-kernels and g-CNNs.

The main novelty of the DGCL is that our architecture changes the discrete parameter learning problem into a parameter sampling problem of the GMM. Therefore, the proposed approach does not rely on the format of the input data. Thus, the DGCL is also valid for feature learning on other irregular input data, such as text data and 3D shape data.

\begin{figure}
\vspace{-0.1in}
\centering
\includegraphics[width=8cm]{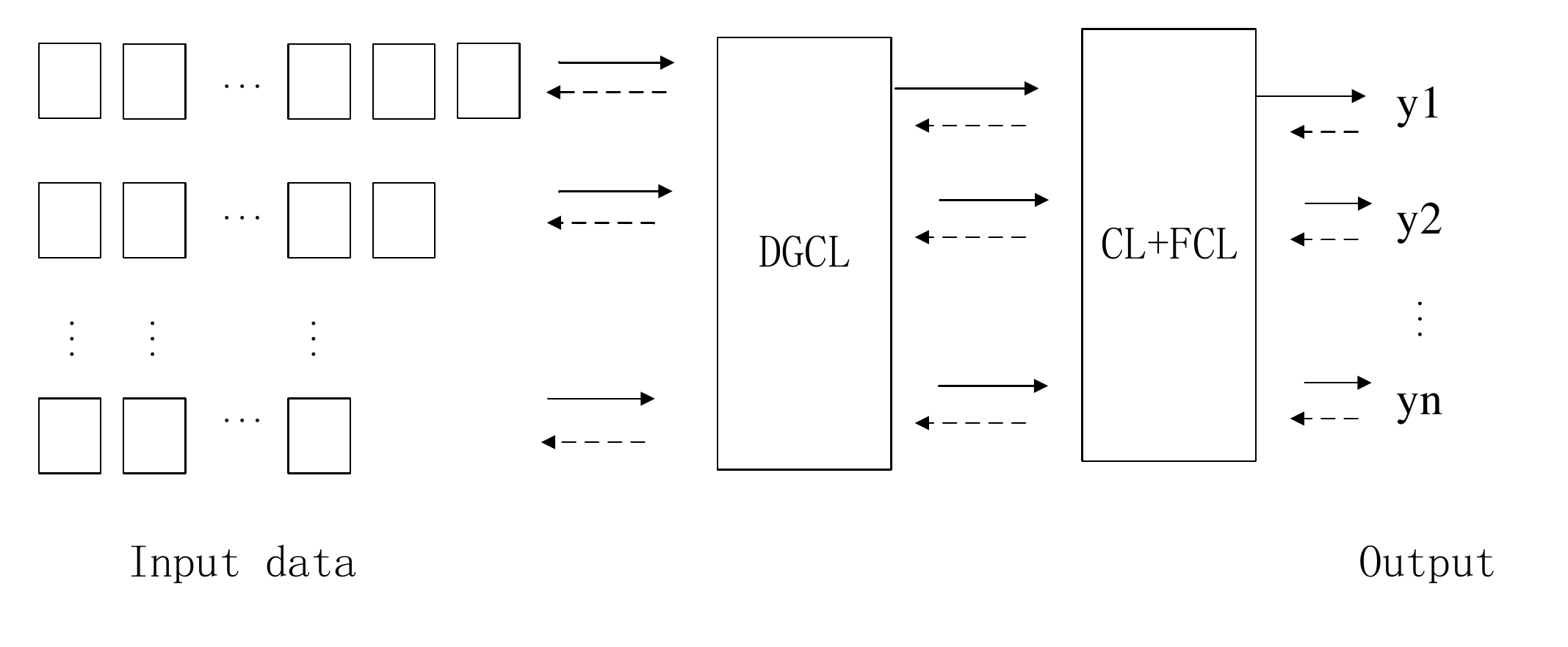}
\vspace{-0.1in}
\caption{the common model of DGCL}\label{fig:common}
\vspace{-0.2in}
\end{figure}

As shown in Fig.~\ref{fig:common}, we represent arbitrary data in matrix form. Then, the matrix serves as the input of the DGCL, and the output of the DGCL can serve as the input of a standard CL or a fully connected layer (FCL). Solid lines denote forward-propagation processes of networks and dotted lines represent back-propagation processes of networks. Therefore, DGCL can be combined with an arbitrary CNN to handle arbitrary regular and irregular data.

\bibliography{AAAI_18}

\end{document}